\documentclass[a4paper]{article}

\usepackage{INTERSPEECH2018}
\usepackage{url}
\usepackage{comment}
\usepackage[normalem]{ulem}

\usepackage{tabularx}
\usepackage{multirow}
\usepackage{microtype}

\newcolumntype{C}{>{\centering\arraybackslash}X}
\newcolumntype{L}{>{\raggedright\arraybackslash}X}

\newcolumntype{b}{X}
\newcolumntype{s}{>{\hsize=.2\hsize}X}

\newcolumntype{e}{>{\hsize=.4\hsize}X}

\newcolumntype{f}{>{\hsize=.5\hsize}X}

\title{Low-Resource Speech-to-Text Translation}
\name{Sameer Bansal$^{1}$, Herman Kamper$^{2}$, Karen Livescu$^{3}$, Adam Lopez$^{1}$, Sharon Goldwater$^1$}
\address{
  $^{1}$School of Informatics, University of Edinburgh, UK \\
  $^{2}$Stellenbosch University, South Africa \\
  $^{3}$Toyota Technological Institute at Chicago, USA}
\email{\{sameer.bansal, sgwater, alopez\}@inf.ed.ac.uk, kamperh@sun.ac.za, klivescu@ttic.edu}

\usepackage{cite}

\usepackage[prependcaption,textsize=scriptsize]{todonotes}
\usepackage{xcolor}
\definecolor{mycolor}{HTML}{FF6600}

\begin{document}

\maketitle
\begin{abstract}
Speech-to-text translation has many potential applications for low-resource languages, but the typical approach of cascading speech recognition with machine translation is often impossible, since the transcripts needed to train a speech recognizer are usually not available for low-resource languages. Recent work has found that neural encoder-decoder models can learn to directly translate foreign speech in high-resource scenarios, without the need for intermediate transcription. We investigate whether this approach also works in settings where both data and computation are limited. To make the approach efficient, we make several architectural changes, including a change from character-level to word-level decoding. We find that this choice yields crucial speed improvements that allow us to train with fewer computational resources, yet still performs well on frequent words.
We explore models trained on between 20 and 160 hours of data, and find that although models trained on less data have considerably lower BLEU scores, they can still predict words with relatively high precision and recall---around 50\% for a model trained on 50 hours of data, versus around 60\% for the full 160 hour model.
Thus, they may still be useful for some low-resource scenarios.

\end{abstract}
\noindent\textbf{Index Terms}: speech translation, low-resource speech processing, encoder-decoder models

\section{Introduction}

Conventional systems for speech-to-text translation~\cite{waibel+fugen_ieee0} typically pipeline automatic speech recognition and machine translation, and since both of these applications require large training sets, these systems are available for only a tiny fraction of the world's highest-resource languages.
But speech-to-text translation could be especially valuable in low-resource scenarios, for example in documentation of unwritten or endangered languages~\cite{besacier2006towards,martin2015utterance,adams2016learning1,adams2016learning,anastasopoulos-chiang:2017:W17-01}; or in crisis relief, where emergency workers might need to respond to calls or requests in a foreign language~\cite{munro2010}.
These applications have motivated recent research on low-resource speech translation trained on a (potentially) cheaper resource: speech paired with its translation, with no intermediate transcriptions.\footnote{There is also recent work~\cite{antonis+tied+naacl18,weiss2017sequence,alexandre+audiobooks} using multitask learning to learn {\em both} translation and transcription models, showing improvements on the individual tasks. We focus here on the scenario without transcriptions.}
Initial work studied speech-to-text alignment without translation~\cite{duong2015attentional,antonios+2016+emnlp+alignment}, or focused on translating a few keywords using heuristic methods with just a few hours of training data~\cite{Anastasopoulos2017spoken,bansal2017towards}.

Recently, recurrent encoder-decoder models have been used to develop end-to-end speech-to-text translation models, which have been tested in high-resource settings on synthesized speech~\cite{berard+etal_nipsworkshop16}, audiobooks~\cite{kocabiyikoglu2018augmenting,alexandre+audiobooks}, and a large dataset of conversational telephone speech~\cite{weiss2017sequence}.
So far, these neural models have been shown to produce high-quality translations with substantial resources---typically more than 100 hours of translated audio, from which models are trained for many days on multiple GPUs. But in our scenarios of interest, we expect to have less data, less time, and fewer computational resources. How will neural models perform in such low-resource settings?

In this paper we perform an extensive investigation of the effects of training end-to-end speech-to-text translation models with limited resources.
We implement a model inspired by the state-of-the-art architecture of Weiss et al.~\cite{weiss2017sequence}, but modify it to permit training in reasonable time on a single GPU (\textsection\ref{sec:model_details}). The biggest change, compared to~\cite{weiss2017sequence} (and also~\cite{antonis+tied+naacl18,alexandre+audiobooks}),
is to use word-level decoding instead of character-level. We show that word-level models can be trained much faster than character-level models and still obtain comparable precision and recall; the tradeoff for this speed improvement is that they struggle to translate infrequent word types, leading to a drop in overall accuracy as measured by BLEU and METEOR.

We investigate the model's performance as we gradually reduce training data from the full 160 hours to as little as 20 hours.
With only 50 hours of training data, our model still produces accurate translations for short utterances, with precision and recall around 50\%.
Although translation quality is much worse with 20 hours of data, precision and recall are still around 30\%, which may be useful in low-resource applications.
The 50-hour model trains in less than three days on a single GPU.

\section{Speech-to-Text Model}
\label{sec:model_details}
Following Weiss et al.~\cite{weiss2017sequence}, we combine convolutional neural network (CNN) and recurrent neural network (RNN) components to build an encoder-decoder model with attention, but we modify the system (Table~\ref{tab:model_diff}) so that we can train even our larger models 
in 3-5 days on a single GPU.

\begin{table}
  \begin{center}
  \begin{tabularx}{\linewidth}{Lff}
    \toprule
     & \multicolumn{1}{c}{\bf Weiss et al.} & \multicolumn{1}{c}{\bf Our model}\\
     \midrule
     speech features & 240 dim & 80 dim \\
     conv LSTM~\cite{xingjian2015convolutional} & yes & no \\
     decoder & characters & words \\
     asynchronous SGD & yes & no \\
     L2 penalty & $10^{-6}$ & $10^{-4}$ \\
     Gaussian weight noise & yes & no \\
     number of model replicas & 10 & 1 \\
    \bottomrule
  \end{tabularx}
  \end{center}\vspace{-1em}
  \caption{Comparison of Weiss et al.~\cite{weiss2017sequence} and our model.\vspace{-2em}}
  \label{tab:model_diff}
\end{table}

\subsection{Speech encoder} 
\label{sub:speech_encoder}

Raw speech input is converted to Mel filterbank features computed every 10ms.
Weiss et al.'s~\cite{weiss2017sequence} model uses 240-dimensional input speech features, consisting of 80 filterbanks stacked with delta and delta-deltas. We use only 80-dimensional filterbank features. In preliminary experiments, we did not find much difference between 40, 40+deltas and 80 dimensions.

The filterbank features are fed into a stack of two CNN layers with rectified linear unit (ReLU) activations~\cite{nair2010rectified}, each with 64 filters (compared to 32 used in Weiss et al.) of shape 3$\times$3 along time and frequency, and a stride of 2$\times$2. Striding reduces the sequence length by a factor of 4, which is important for reducing computation in the subsequent RNN layers.

At training time we use bucketing---80 buckets, with width increments of 25 frames---and padding of speech data. The utterances in the training set vary in length from 2 to 30 seconds; those longer than 20 seconds (80$\times$25 frames) are truncated. We select up to 64 utterances from a bucket to create a mini-batch.

The output of the CNN layers is fed into three stacked bi-directional long short-term memory (LSTM)~\cite{hochreiter+lstm} layers, with 256-dimensional hidden states in each direction. Since the RNN operations are the main bottleneck, for initial hyper-parameter tuning we used uni-directional LSTMs with 300-dimensional hidden states. 
We then switch to bi-directional LSTM encoders to generate the final results.


\subsection{English decoder} 
\label{sub:english_decoder}
We use a word-level decoder, whereas Weiss et al.~\cite{weiss2017sequence} used a character-level decoder. Since there are roughly five times as many characters as words, this greatly reduces sequence length, which speeds training for each individual utterance and allows us to use larger mini-batch sizes.

The English words are fed into an embedding layer followed by a stack of three uni-directional LSTM layers.
We implement attention using the {\em global attentional model} with {\em general} score function and {\em input-feeding}, as described in~\cite{luong2015effective}.
We use beam decoding with a beam size of 8.


\subsection{Optimizer} 
\label{sub:optimizer}
We use cross-entropy as the loss function. 
We regularize with dropout~\cite{srivastava+dropout}, with a ratio of $0.5$ over the embedding and the LSTM layers~\cite{Gal2015Theoretically}, and an L2 penalty of $0.0001$. We use a teacher-forcing~\cite{williams+teacher_forcing} ratio of 0.8.
Our code is implemented in Chainer~\cite{chainer_learningsys2015}.{\footnote{Code available at: \url{https://github.com/0xSameer/speech2text/tree/seq2seq}}}
Weiss et al.'s~\cite{weiss2017sequence} model is trained using asynchronous stochastic gradient descent (ASGD) across 10 replicas; we train using a single model copy.
Although Weiss et al.'s model benefited greatly from adding Gaussian noise to the weights during training (personal communication), we were unable to replicate this benefit and did not use noise injection.


\section{Experiments}

\subsection{Experimental setup}
We use the Fisher Spanish speech dataset~\cite{LDC2010S01}:
a multispeaker corpus of telephone calls in a variety of Spanish dialects recorded in realistic noise conditions. The English translations were collected through crowdsourcing, as described in \cite{post2013improved}, and are used to train all models.
There are four English references per utterance for the development and test sets, and one per utterance for the training set.
We only use one of the two development sets ({\em dev}, not {\em dev2}).
The training set comprises 160 hours of speech, split into 140K utterances; the development and test sets have about 4.5 hours of speech split into 4K utterances each.
We lower-case and remove punctuation from the English translations and tokenize the text using NLTK~\cite{loper+nltk}.{\footnote{\url{http://www.nltk.org/api/nltk.tokenize.html#nltk.tokenize.word_tokenize}}} 
This gives about 17K training word types and 1.5M tokens. There are about 300 out-of-vocabulary (OOV) word types (400 tokens) in the {dev} set, out of 40K tokens.

We first train a model using the entire 160 hours of labeled training data. To understand the impact of training data size on translation quality, we further train models using smaller subsets: 80, 50, 25, 20 hours of data, selected at random, from the entire training data. We use the same set of hyper-parameters---tuned for 160hrs---for all models.
With these model parameters and training setup, we are able to train an epoch---a complete pass through the entire 160 hours of training data---in about $2$ hours on a single Titan X (or equivalent) GPU.

\subsection{Evaluation} 
\label{sub:evaluation}
In order to understand different aspects of model behavior, we evaluate with several metrics: BLEU~\cite{papineni+bleu}, METEOR~\cite{lavie+meteor}, and unigram precision/recall on the Fisher {\em dev} set, using all 4 human references.\footnote{BLEU and precision are computed using {\tt multi-bleu.pl} from the MOSES toolkit~\cite{koehn2007moses}, which computes 4-gram BLEU. METEOR score and recall are computed using the script from \url{http://www.cs.cmu.edu/~alavie/METEOR/}.} BLEU measures how well a predicted translation matches a set of references based on a modified $n$-gram precision; it does not compute recall, and instead uses a {\em brevity penalty} to account for mismatch in reference and predicted lengths. METEOR computes both precision and recall and combines them via a harmonic mean, with greater weight given to recall.
The final score is computed using a set of parameters, tuned for individual languages to correlate well with human judgments.

Whereas BLEU looks only for exact token matches between a prediction and set of references, METEOR also takes into account {\em stem}, {\em synonym}, and {\em paraphrase} matches. These four categories are weighted (by default) 1.0, 0.6, 0.8 and 0.6, respectively. For example, with these weights, a prediction of ``eat'' will score a recall of 1 against reference ``eat'' and 0.8 against reference ``feed'', a synonym match.
METEOR can therefore be considered a more semantic measure. In low-resource settings, inexact translations that capture the semantics of an utterance are still useful. We use default settings and configuration files provided by the METEOR script for English. For comparison, we also provide human-level BLEU and METEOR scores by comparing one reference against the remaining 3.

In low-resource settings, BLEU scores may be very low and therefore difficult to interpret, but a model might still be able to predict (potentially important) keywords, which could be useful for cross-lingual applications. So, we also report word-level unigram precision using the BLEU script: if a predicted token is present in any of the 4 reference translations, it is considered a {\em True Positive}; otherwise it is a {\em False Positive}.
Unigram recall is computed using the METEOR script, which includes {\em stem}, {\em synonym}, and {\em paraphrase} level matches. We also compute recall for {\em exact} matches only, by setting the METEOR weights to 1.0, 0.0, 0.0 and 0.0.

\subsection{Results and discussion} 
\label{sec:results}
\label{sec:discussion}

\begin{figure}[t]
  \centering
  \includegraphics[width=\linewidth]{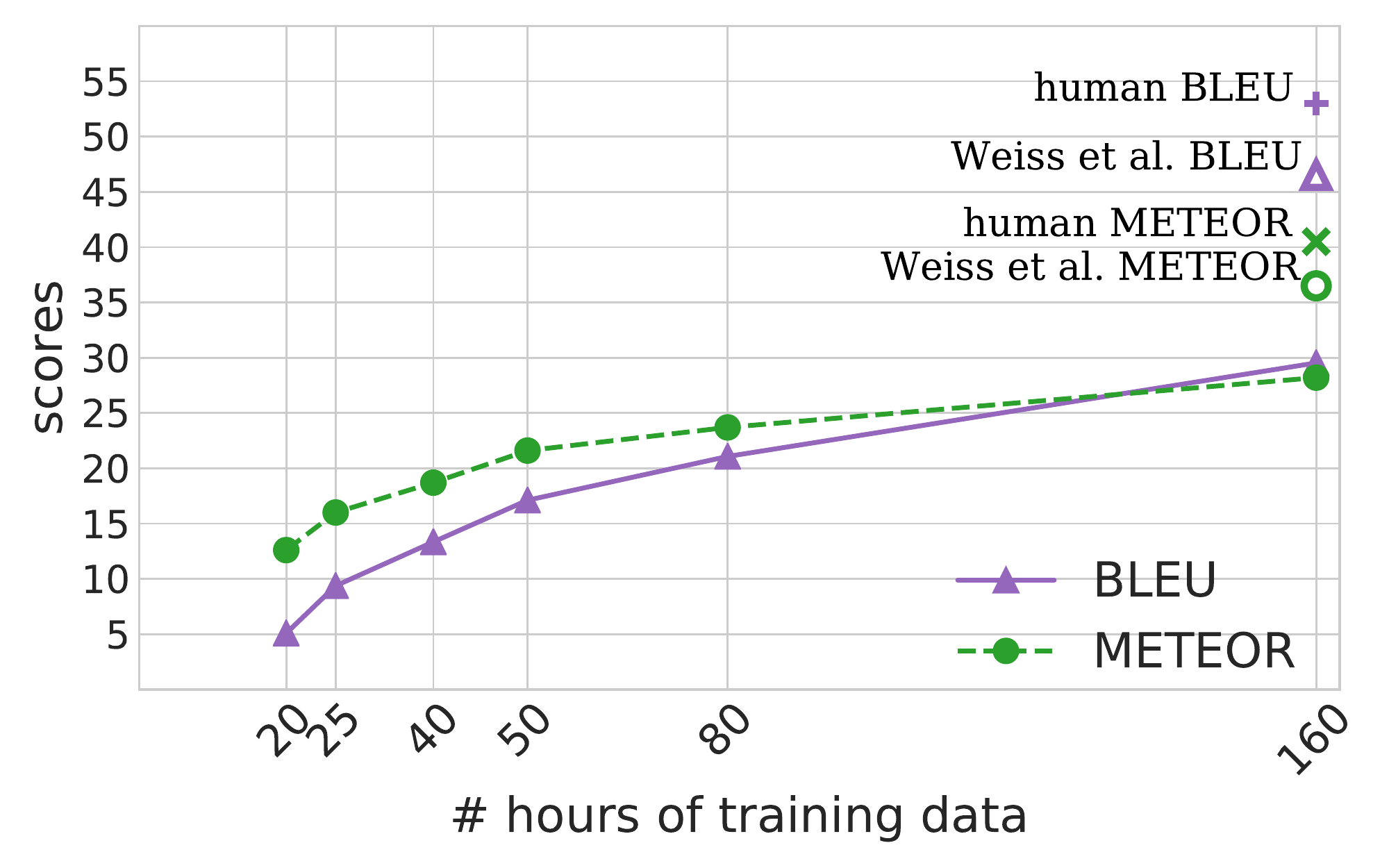}
  \caption{Fisher dev set BLEU/METEOR results.}
  \label{fig:bleu_meteor}
\end{figure}

\begin{figure}[t]
  \centering
  \includegraphics[width=\linewidth]{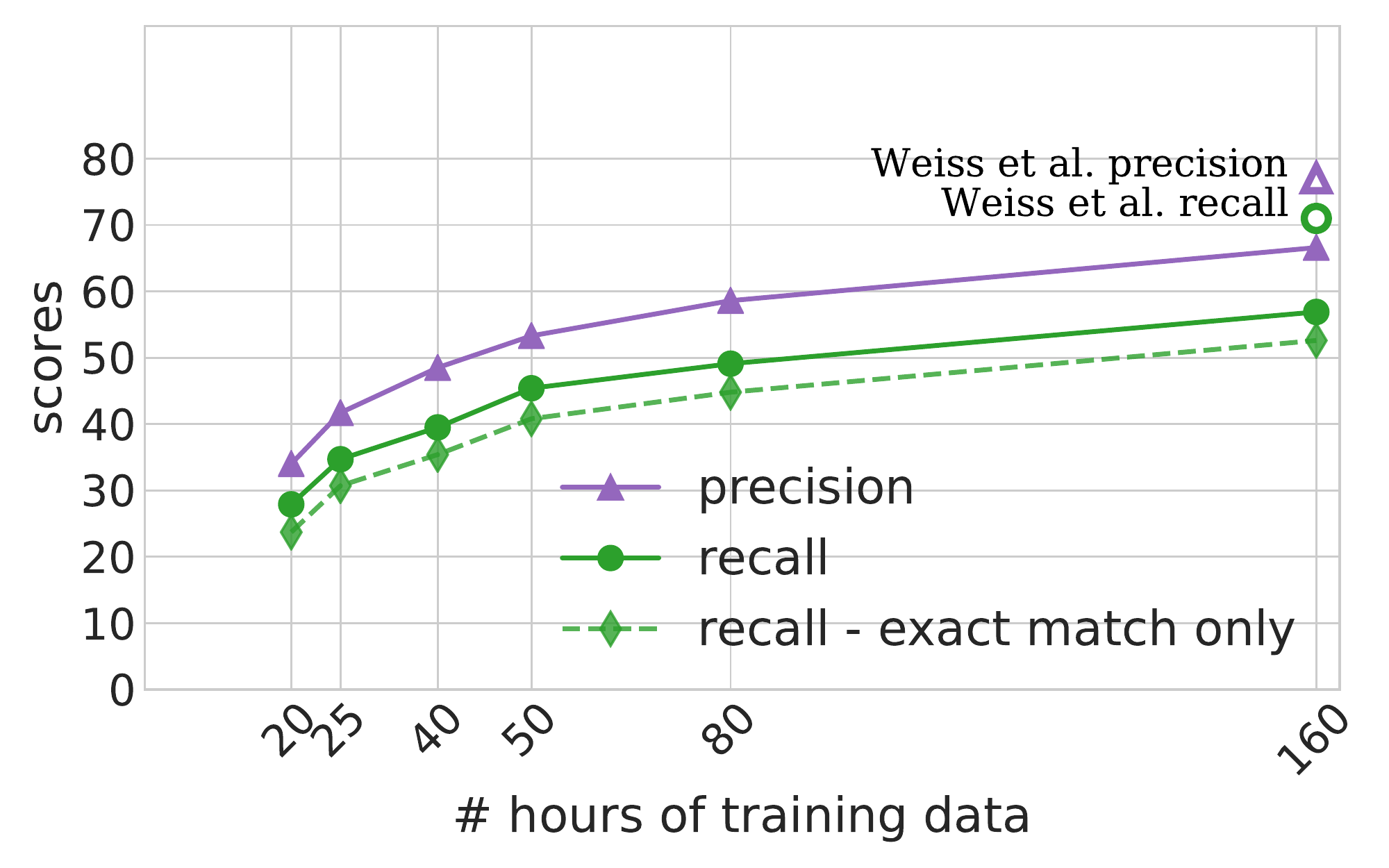}
  \caption{Fisher dev set precision/recall results.}
  \label{fig:precision_recall}
\end{figure}

Figures~\ref{fig:bleu_meteor} and~\ref{fig:precision_recall} show the BLEU, METEOR, and precision/recall scores on the {dev} set for each model as we change the amount of training data.
Table~\ref{tab:bleu_test} shows the BLEU scores on the {\em test} set.

\begin{figure}[t]
  \centering
  \includegraphics[width=\linewidth]{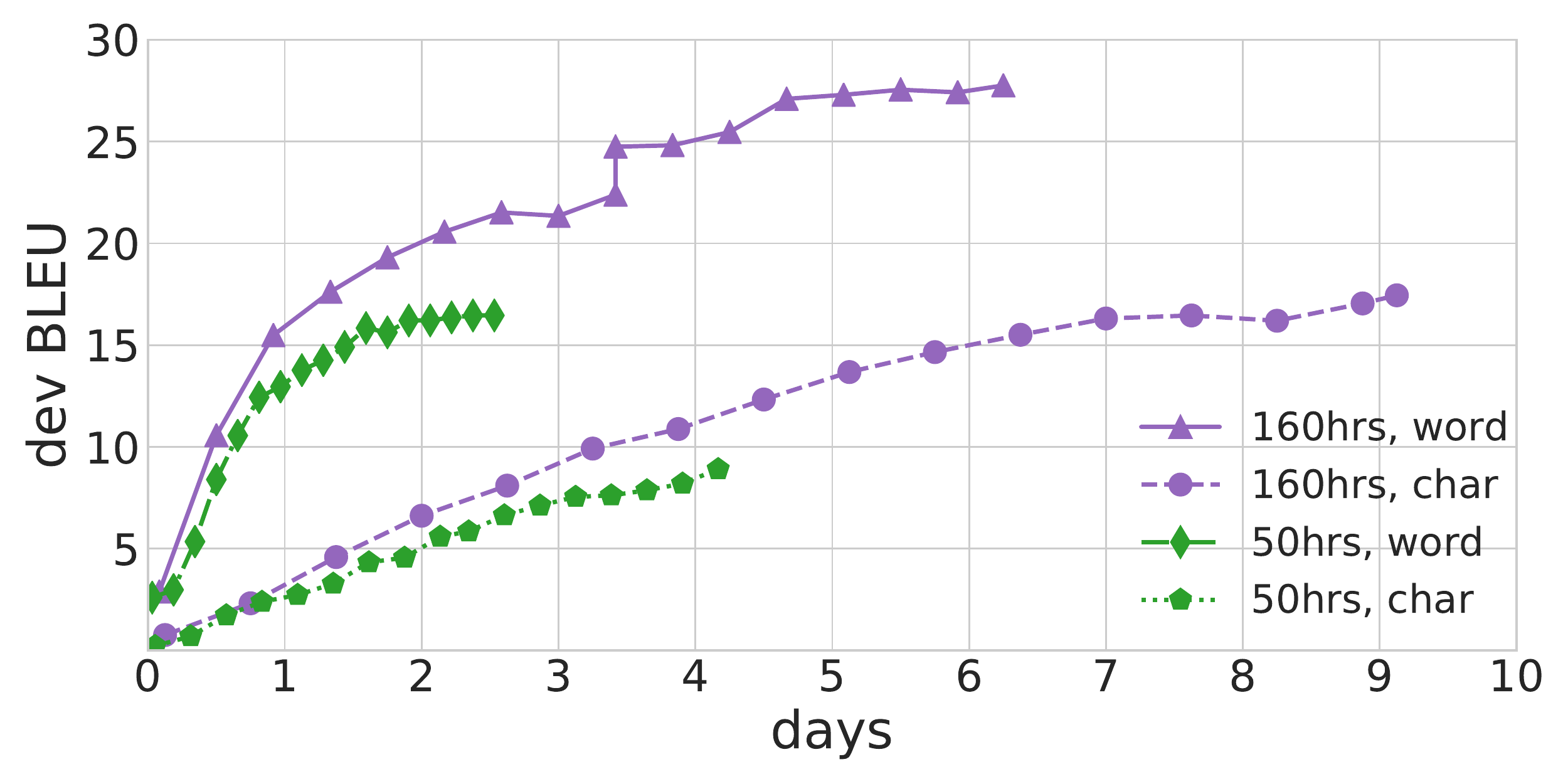}
  \caption{Performance vs.~training time for the word vs.~character decoders. Each marker denotes 5 epochs.}
  \label{fig:bleu_word_vs_char_epochs_days}
\end{figure}

\vspace{.1in}
\noindent {\bf Word vs.~character models.} Weiss et al.'s~\cite{weiss2017sequence} character-level model achieves close to human performance.\footnote{The human scores are computed using 3 references; BLEU scores for all models are 2-4 points lower when using 3 instead of 4 references.}
Our word-level model, trained on 160 hours of data, converges to a BLEU score of 29.5 in about five days, a much lower score than Weiss et al.'s.
One reason for this discrepancy is our different architecture and training setup (\textsection\ref{sec:model_details} and Table~\ref{tab:model_diff}), which allowed us to train our models on our available computing resources.\footnote{Since our paper was submitted, \cite{antonis+tied+naacl18} reported results on 20 hours of Spanish-English data for several multitask (translation/transcription) models and a baseline speech-to-text model. They used a character-level decoder and a different corpus (CALLHOME). They did not report detailed word-level BLEU scores, but said they were ``between 7 and 10'' for all models.}

Training the character-level model takes nearly twice as long per epoch (4 hours for 160hrs of data) as the word-level model (2 hours).
Our character-level model also has much smaller performance gains per epoch. To speed up character-level model training, we truncate utterances longer than 15 seconds (20 seconds for word models), reducing training time to 3 hours per epoch.
Figure~\ref{fig:bleu_word_vs_char_epochs_days} compares character-level models trained in this way to word-level models for two training set sizes.

\begin{table}
\vspace{1em}
  \begin{center}
  \begin{tabularx}{0.9\linewidth}{CCCCCCC}
    \toprule
      {\bf W} & {\bf 160h} & {\bf 80h} & {\bf 50h} & {\bf 40h} & {\bf 25h} & {\bf 20h}\\
     \midrule
     47.3 & 29.4 & 21.4 & 18.2 & 13.6 & 8.9 & 5.3 \\
  \bottomrule
  \end{tabularx}
  \end{center}
  \caption{BLEU scores of ({\bf W})eiss et al.'s model and our models on the Fisher test set.
  \vspace{-1em}}
  \label{tab:bleu_test}
\end{table}

\begin{table}[b]
  \begin{center}
  \begin{tabularx}{0.9\linewidth}{Leee}
    \toprule
     & \multicolumn{1}{c}{\bf Rare} & \multicolumn{1}{c}{\bf Medium} & \multicolumn{1}{c}{\bf Frequent}\\
     \midrule
     training types & 12K & 1K & 386    \\
     training tokens & 30K & 56K & 200K \\
     \midrule
{\em Precision (\%):}\\
     Weiss et al.           & 81.1 & 76.3 & 79.6 \\
     {160hrs}               &  87.5 & 69 & 65.7 \\
    \midrule
{\em Recall (\%):}\\
     Weiss et al.           & 24.5 & 65.4 & 78.1 \\
     {160hrs}               & 1.1 & 36.9 & 64.4 \\
    \bottomrule
  \end{tabularx}
  \end{center}
  \caption{Content word frequency vs.~dev precision/recall. {\em Rare} words have $\le$10 tokens per type in the training text; {\em medium} have 25--100 tokens;  {\em frequent} have $\ge$150 tokens.}
  \label{tab:prec_recall_word_freq}
\end{table}

One of the main benefits of character models is their ability to gracefully handle OOV or infrequent words.  On the {dev} set, the Weiss et al.~model predicts about 130 word types that were not seen in training, which helps the model recall 7 OOV tokens out of 400.  This is too small an effect to account for the performance differences, so we also analyze performance for a range of word frequencies.
Table~\ref{tab:prec_recall_word_freq} shows the precision/recall for the 160hrs model and Weiss et al.'s~\cite{weiss2017sequence} model,
for words of different frequencies.
We only consider content word types---words that are more than five characters long and are not in the NLTK stopword lists.
The word-level model recall drops rapidly for {\em medium} frequency words, and for {\em rare} word types it has almost 0\% recall. From this, we see that the main benefit of the character-level model is in handling of rare words, rather than previously unseen words.

\begin{figure}[t]
  \centering
  \includegraphics[width=\linewidth]{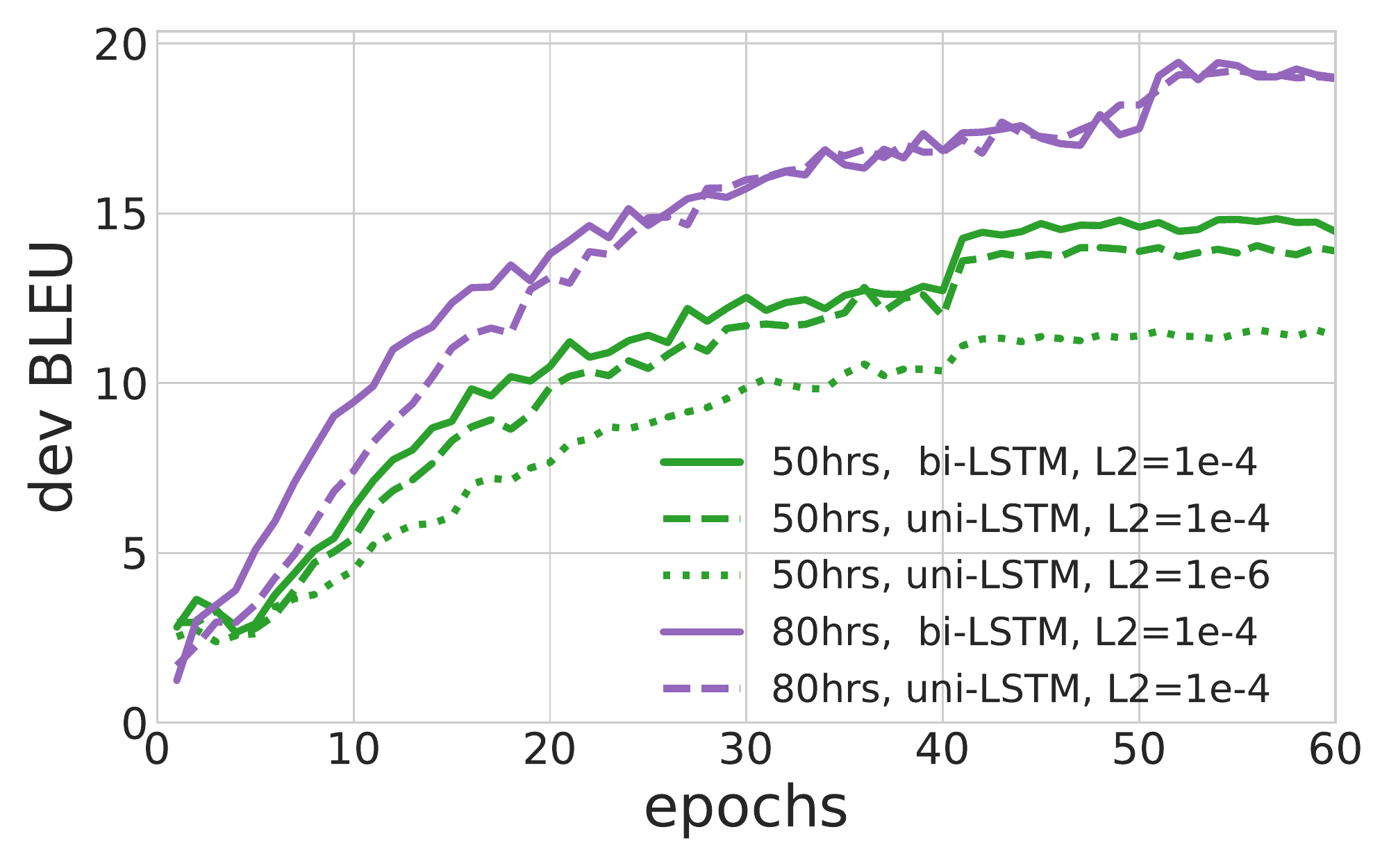}
  \caption{Performance comparison: uni-directional vs.~bi-directional encoders, L2 loss penalties.}
  \label{fig:bleu_vs_modelparams}
\end{figure}

\vspace{.1in}
\noindent {\bf Training considerations.}
Regularization parameters are critical to model performance. Figure~\ref{fig:bleu_vs_modelparams} shows that increasing the L2 weight decay to a rate of $10^{-4}$ from $10^{-6}$ improved BLEU by about 2 points.
Even though we use a high L2 penalty and dropout ratio, the models can overfit;
the training loss continues to decrease, and we use early stopping based on {dev} set BLEU.

We also tried using batch normalization~\cite{ioffe+batchnorm+arxiv_2015} at each CNN layer, and layer normalization~\cite{ba+layernorm+arxiv+2016} at each LSTM layer; but these did not have any noticeable impact on training performance.

\vspace{.1in}
\noindent {\bf Other design considerations.}  Using a bi-directional LSTM encoder does not seem to have an effect for the largest training sets, but improves BLEU by about one point in the $\le$50 hour cases
(Figure~\ref{fig:bleu_vs_modelparams}). This comes at a training time cost: bi-directional encoder models are almost $50\%$ slower to train per epoch.

\begin{figure}[t]
  \centering
  \includegraphics[width=\linewidth]{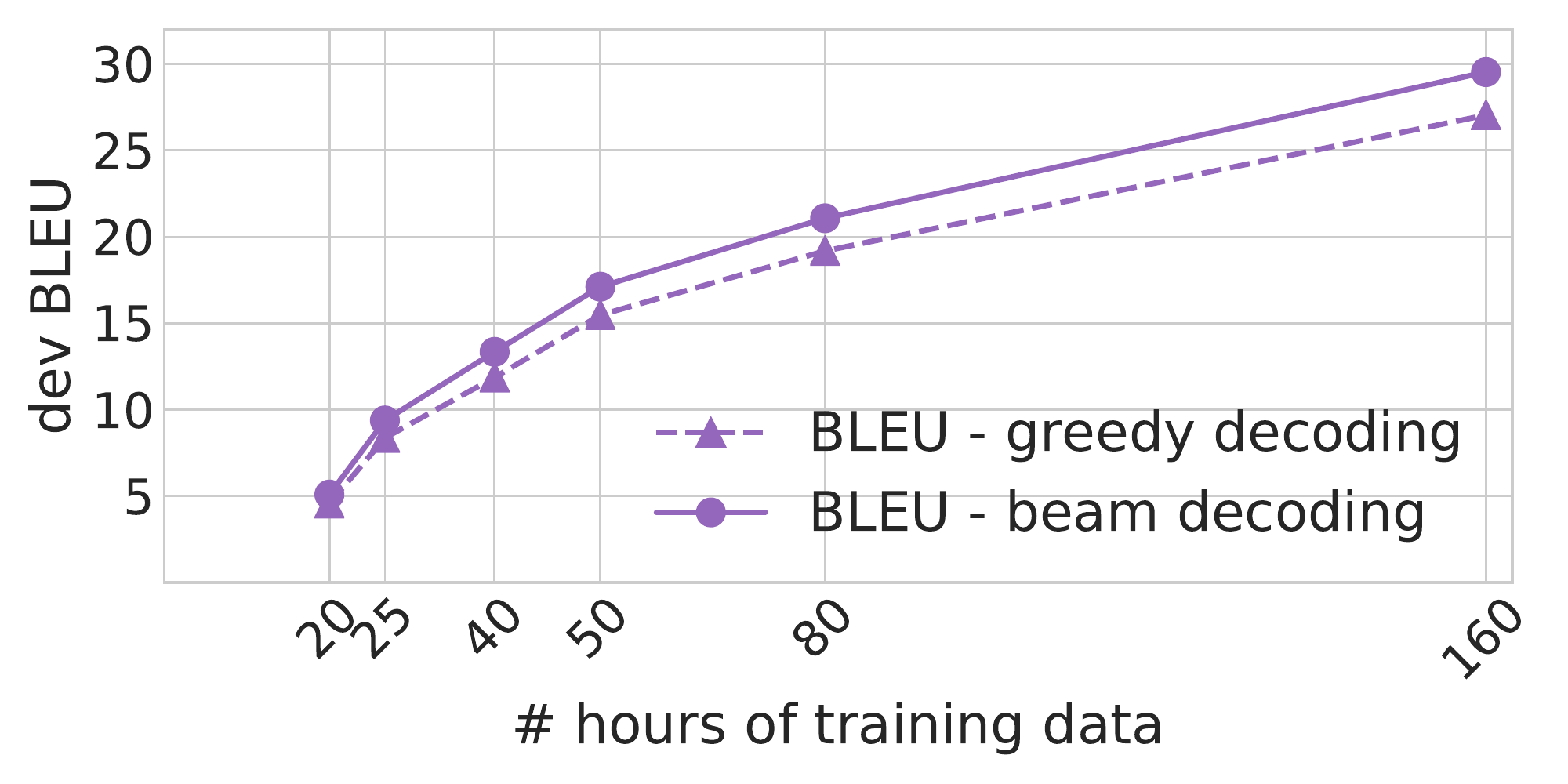}
  \caption{Performance comparison: Greedy vs.~beam decoding.}
  \label{fig:bleu_beam}
\end{figure}

Figure~\ref{fig:bleu_beam} shows the improvement in BLEU scores by using beam decoding, over greedy decoding. Beam decoding always helps, but has a larger effect with more training data.

\begin{table}
\vspace{1em}
\begin{center}
  \footnotesize
  \begin{tabularx}{\linewidth}{l@{\hspace{2mm}}b}
    \toprule

    {\bf model} & {\bf translations} \\
    \midrule
    {\it Ref}  &  so no yes but there are people who do get bothered a lot  \\
     \noalign{\vskip 1.5mm}
    W &  \underline{so no yes} \underline{there are people} that \underline{do} \underline{bother} \underline{a lot}\\
    {160h}   &  \underline{so no} if \underline{people} \underline{are} \underline{bother}ing \underline{a lot} \\
    {80h}   &  \underline{so} if \underline{there are people} that are \underline{bother}ing \underline{a lot} \\
    {50h}   &  \underline{so no yes} that's why it \underline{bother}s me \underline{a lot} \\
    {40h}   &  \underline{so} if you think that it's like \underline{a lot} \\
    {25h}  &  \underline{so} i don't know if \underline{people} who are \underline{bother} me much \\
    {20h}  &  \underline{so} if you have a car you can do it \underline{a lot}\\
    \midrule
    {\it Ref}  &  greetings ah my name is jenny and i'm calling from new york \\
     \noalign{\vskip 1.5mm}
    W  & hi \underline{ah my name is jenny} \underline{i'm calling from new york} \\
    {160h} & hi \underline{ah my name is jenny} \underline{i'm calling from new york} \\
    {80h} & good \underline{ah my name is jenny} \underline{i'm calling from new york} \\
    {50h} & good \underline{ah my name is jenny} \underline{calling from new york} \\
    {40h}   &  well \underline{ah} \underline{i'm calling from from new york} \\
    {25h}  &  good \underline{ah my name is} peruvian \underline{i'm calling from new york} \\
    {20h}  &  good \underline{ah my name is jenny} \\
  \bottomrule
  \end{tabularx}
  \end{center}
  \caption{Example translations of ({\bf W})eiss et al.'s model and our models on dev set utterances, with stem-level $n$-gram matches to the reference sentence underlined. \vspace{-1em}}
  \label{tab:sample_translations}
\end{table}

\vspace{.1in}
\noindent {\bf Exact vs.~semantic matches.}
Figure~\ref{fig:bleu_meteor} shows that the gap in METEOR scores between our models and Weiss et al.'s is much smaller than the gap in BLEU, and METEOR degrades more slowly as training data is reduced.
This suggests that although our models are much worse
at predicting the exact words in the reference translations, they often predict near-exact matches.

Table~\ref{tab:sample_translations} shows some example predictions.
As expected from the BLEU/METEOR results, the translations trained on more than 50 hours are fairly good, though they may contain different forms of the content words than are in the reference (e.g.~{\em bothering/bothers} vs.~{\em bothered}).
The models trained on less data are clearly worse:
they usually get some words right, which could be useful for keyword spotting or topic modeling in low-resource settings, but in some cases (as in the first example) the correctly predicted words do not carry much of the meaning.

\vspace{.1in}
\noindent {\bf Neural models in the extreme low-data setting.}  One may wonder whether, for very low-data settings, neural models still outperform older non-neural models at all.  While we did not directly compare to a non-neural model, one indication is that,
in the lowest-data settings, BLEU and METEOR are much worse but unigram precision and recall are still in the 25-35\% range. These results compare very favorably to the 2-10\% precision/recall reported by \cite{bansal2017towards}, who used a heuristic speech-to-text translation system trained on the CALLHOME corpus (also about 20 hours of Spanish conversational telephone speech with English text translations). So, it appears that even in a very low-resource setting with a model that is not state-of-the-art, the neural approach significantly outperforms previous non-neural models.

\section{Conclusion}

We performed a thorough analysis of a neural end-to-end speech-to-text translation model, with a specific focus on how performance is affected when using limited computational resources and limited amounts of data.
We also showed the effects of a number of architectural design decisions using several performance metrics.
While word-level models fall behind previously proposed character-level models when trained on around 160 hours of translated speech, our word-level models can be trained much faster and give reasonable performance on smaller training sets.
Although translation quality drops, models trained on only 20 hours of translated speech achieve precision and recall of around 30\% for content words.
This could still be useful in search applications in severely low-resource scenarios.
We believe that our extensive analyses in this work will contribute to better decision-making for architectural choices in computation- and data-limited settings.

In future work we aim to consider sub-word modelling~\cite{sennrich-haddow-birch:2016:P16-12}, which could balance the trade-off between training costs and translation performance.
In addition, we plan to try speech features that are targeted to low-resource multi-speaker settings~\cite{kamper2015unsupervised,kamper+2016+arxiv+fullsegmental} and speaker normalization \cite{zeghidour+etal_interspeech16}.


\section{Acknowledgements}
We thank Ron Weiss and Jan Chorowski for sharing their translation output, and Kenneth Heafield for giving access to GPUs. This work was supported in part by a James S. McDonnell Foundation Scholar Award and a Google faculty research award.

\bibliographystyle{IEEEtran}
\bibliography{sp2t}

\end{document}